\documentclass{article}

     \usepackage[nonatbib]{neurips_2020}

\usepackage[utf8]{inputenc} 
\usepackage[T1]{fontenc}    
\usepackage{hyperref}       
\usepackage{url}            
\usepackage{booktabs}       
\usepackage{amsfonts}       
\usepackage{nicefrac}       
\usepackage{microtype}      
\usepackage{graphicx}
\usepackage{cite}
\usepackage{xurl}
\usepackage{subfig}

\title{TransDocs: Optical Character Recognition with word to word translation}

\author{%
  David S.~Hippocampus\thanks{Use footnote for providing further information
    about author (webpage, alternative address)---\emph{not} for acknowledging
    funding agencies.} \\
  Department of Computer Science\\
  Cranberry-Lemon University\\
  Pittsburgh, PA 15213 \\
  \texttt{hippo@cs.cranberry-lemon.edu} \\
}

\begin{document}

\maketitle

\begin{abstract}
    While OCR has been used in various applications, its output is not always accurate, leading to misfit words. This research work focuses on improving the optical character recognition (OCR) with ML techniques with integration of OCR with long short-term memory (LSTM) based sequence to sequence deep learning models to perform document translation. This work is based on ANKI dataset for English to Spanish translation. In this work, I have shown comparative study for pre-trained OCR while using deep learning model using LSTM-based seq2seq architecture with attention for machine translation. End-to-end performance of the model has been expressed in BLEU-4 score. This research paper is aimed at researchers and practitioners interested in OCR and its applications in document translation.
\end{abstract}

\section{Introduction}
The Optical character recognition (OCR) has been widely used in various application scenarios, such as online education, toll gates operation, industrial automation and even robotics. OCR technique basically aims to recognize text from wide variety of images from scanned paper documents to text in wild. The technique has a long research history and has been recently achieved very good results in terms of accuracy.\cite{DBLP:journals/corr/abs-1904-01906} In this project, I'm focusing on a new application of OCR technique by integrating an Long short-term memory (LSTM) based sequence to sequence deep learning model\cite{8543868, Uppala-2021-129108} to perform document translation. This is a challenging pipeline to perform as the output from OCR model is not always correct and can produce misfits like 'C0de' for the original word 'Code'. 

I have trained a LSTM based model which can take inputs from OCR outputs which can have misfit words and the model should give word by word translation and show translation corresponding to the input given from the OCR output. I have used ANKI dataset\cite{ANKI} for training the translation model (English to Spanish) to train the baseline. 

To further enhance my performance on the misfit readings from the OCR, I have augmented the data by using OCR-output with ground-truth from the ANKI's converted pair. The dataset for the OCR testing and generating ground truth for the translation, I have used TextRecognitionDataGenerator \cite{textdatagen, kar2023keydetect} which is a well known dataset generator for OCR model training. The text data generator randomly chooses words from a dictionary of a specific language or an input file or as in my case, choose from a translation pair (English - Spanish pair). Then an image of those words has been generated by using different font, background, and modifications (skewing, blurring, background and distortion) as specified, along with the ground truth. Once the images are generated from the text generator, images are fed into the OCR model and the prediction are extracted. 

\begin{figure}[!h]
  \centering
  \includegraphics[width=0.7\textwidth]{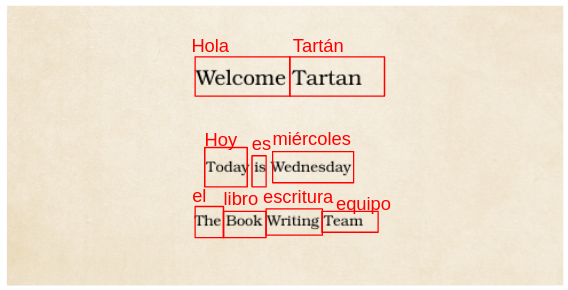}
  \caption{Example of output of the TransDocs pipeline for English to Spanish translation.}
  \label{fig:demo_figure}
\end{figure}

The predicted text is then mapped with the original text and hence, the translation. The final output of the complete pipeline is the translation of the original image and can be presented as shown in figure \ref{fig:demo_figure}, the result shown is an ideal case and shown to demonstrate the final result. In this project, I have not worked very complete ANKI dataset, but only chose to work on a one language pair because of limited resources and very high training time for seq2seq models.

\section{Background}
OCR is one of the most important challenges in computer vision and many methods have been proposed over the recent years to tackle the problem. The OCR can be understood as a classification task by considering each character of words and classify to the corresponding category. There have been various attempts to improve the technique and one of the best models with high accuracy are Easy OCR \cite{easyocr} and Tesseract \cite{tesseract}. OCR models run in 3 stages as: Line/Word finding, Word recognition and Static Character Classifier. 

Machine translation has been around for a while and has been an active research topic. There are many companies trying to solve the problem in both textual and audio format. In this project, I'm focusing on working with the textual machine translation. There are various different models available for text language translation and have been improved over the years. Some notable models are Google seq2seq \cite{seq2seqgoogle}, Transformer model \cite{vaswani2017attention} and  Facebook FAIR's submission to the WMT19 \cite{facebookseq}. The above stated models are one of the top contributing models have high BLEU score. Higher the BLEU better is the performance of the machine translation.

\section{Related work}

Machine translation is a really interesting problem to learn and solve. I had to reference many resources to learn about the architecture and the approach to be adopted. Before performing the machine translation, I had to choose been the best pre-trained OCR model for reading the text from the images. 

\subsection{Optical Character Recognition}
I had to review the documentation for for EasyOCR and Tesseract. Both the models performed well in over-powered the counter in certain conditions.

\begin{figure}[!h]
  \centering
  \includegraphics[width=0.9\textwidth]{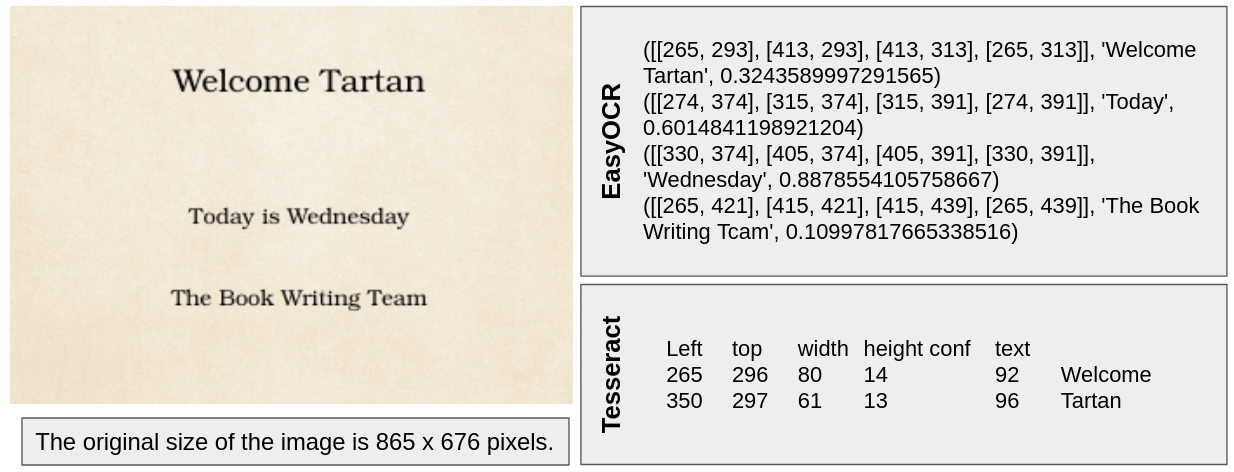}
  \caption{Introduction cover example wth easyOCR and Tesseract model with bounding box information.}
  \label{fig:example1}
\end{figure}

\begin{figure}[!h]
  \centering
  \includegraphics[width=0.9\textwidth]{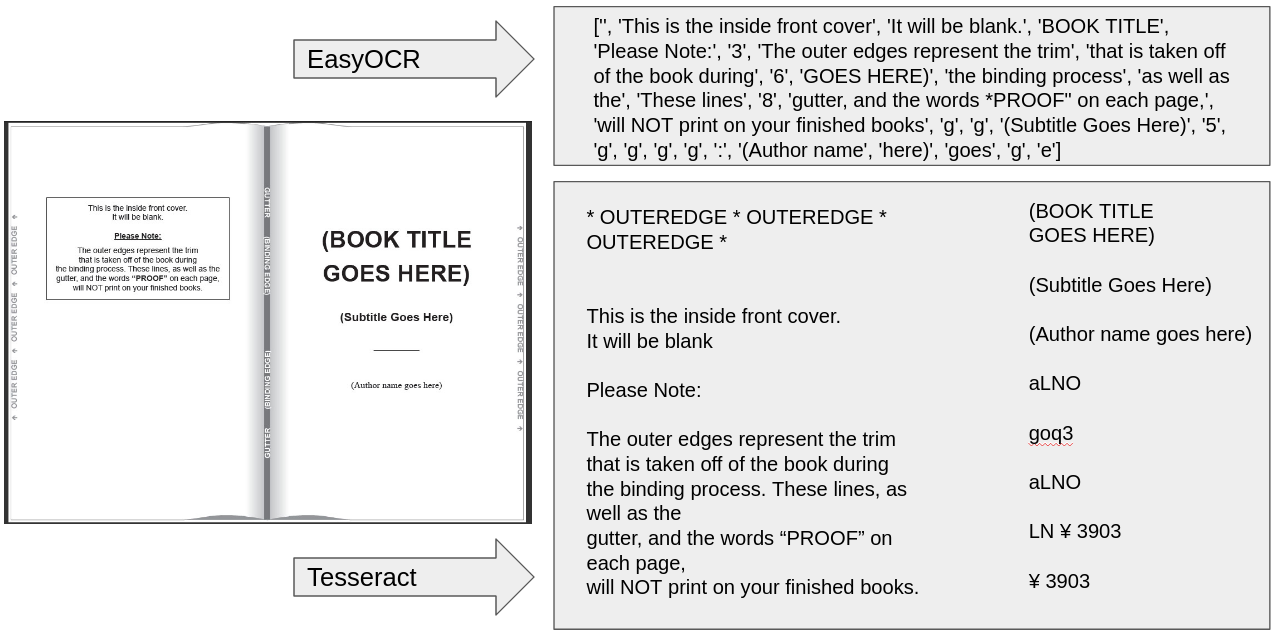}
  \caption{Book title page example wth easyOCR and Tesseract model without bounding box information.}
  \label{fig:example2}
\end{figure}

The results shown in figures \ref{fig:example1} and \ref{fig:example2} are using easyOCR and Tesseract pre-trained models. Predictions were run on 50 images and the two interesting results are shown in the figures. In introduction cover example, we can see that easyOCR is able to outperform Tesseract model, interestingly, tesseract is not able to detect any other words on the image and gives empty strings for some detecting areas (not shown in the figure). This is because of the background/noise in the example. Though tesseract is believed to be very strong model. Whereas, when we compare book title page example, we can see that tesseract shows much detailed results for the example image. It is able to detect and recognize significantly oriented text very effectively. When compared to the result for easyOCR, it is not able to perform well on high varying orientation text. Also, varying text sizes and fonts can also result in degraded results from both models. 

\subsection{Machine translation}
Once I have the predictions from the OCR, I had to learn about the machine learning models currently available. I analysed that the most used models are Sequence to Sequence models with LSTM, GRU and RNN. There are great models published in the past and have a really good BLEU score. Some notable models are Google seq2seq \cite{seq2seqgoogle}, Transformer model \cite{vaswani2017attention, uppala2023learning} and  Facebook FAIR's submission to the WMT19 \cite{facebookseq}. I tried the pre-trained model for FairSeq was able to get good result for perfect inputs and erroneous output for misfit inputs.

\begin{figure}[!h]
  \centering
  \includegraphics[width=0.9\textwidth]{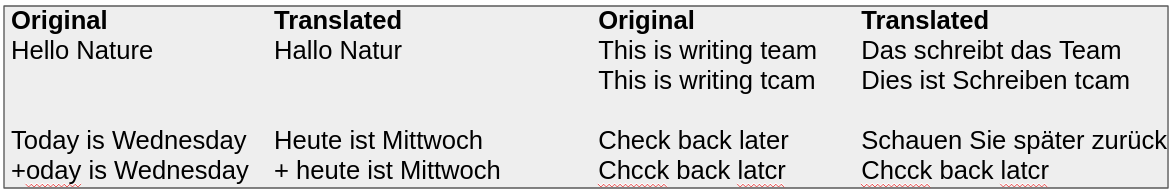}
  \caption{Results using fairseq }
  \label{fig:fairseq}
\end{figure}

In the figure \ref{fig:fairseq}, we can see that the existing models are not able to produce correct machine translation for misfit inputs as compared to the original correct input. I have tried to produce results for both transformer.wmt19.en-de.single\_model and conv.wmt17.en-de and produces similar results.

To better understand the architecture involved in developing machine translation networks, as a start point I revisited the lecture on LSTM which was taken by Prof. Andrej. I visited the official documentation by Keras on 'A ten-minute introduction to sequence-to-sequence learning in Keras'\cite{keras_official}. To explore the attention models, I explored the Bahdanau et al. model \cite{bahdanau2014neural} and Luong et al. model \cite{Luong}. Did thorough comparison between the two attention models for the machine translation and the impact on learning. Later in the models section, I have discussed about the attention model adopted.

\section{Methods/Models}
\subsection{Data pre-processing}
\begin{figure}
    \centering
    \includegraphics[width=0.5\textwidth]{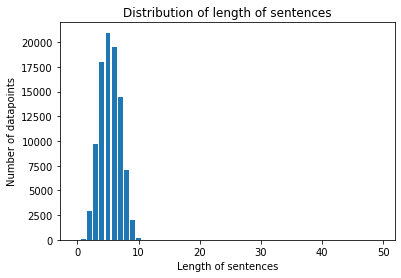}
    \caption{Distribution of number of words in each strings}
    \label{fig:distribution}
\end{figure}

In this project, I have used ANKI dataset (Spanish - English) to train my machine translation model. The dataset is already very well structured and didn't require much pre-processing like remove invalid characters, missing entries or no translation. Once the dataset was read, a few pre-processing had to be done to get it ready for the training phase. The data was converted into lower case and all the punctuation was removed from all the languages. Once the punctuation were removed, I normalized the data and also only kept the alphabets in the strings. Converted the normalized strings into 'UTF-8' format. Just to make it consistent and to remove any extra spaces in the strings, string was split into words and then removed all spaces from the words and empty strings. The list of processed word was joined together with single space between the words. After the data pre-processing, I analysed the number of words in each string and created a bar plot and can be seen in the figure \ref{fig:distribution}. From the graph, we see that all the strings are 10 words or less. But, as the y-axis is a very big scale, there are a smaller values in the range of 10s for strings in with number of words in the the range of 20-50 words per string.

\subsection{Data augmentation for misfit words}
\begin{figure}[!h]
    \centering
    \includegraphics[width=0.9\textwidth]{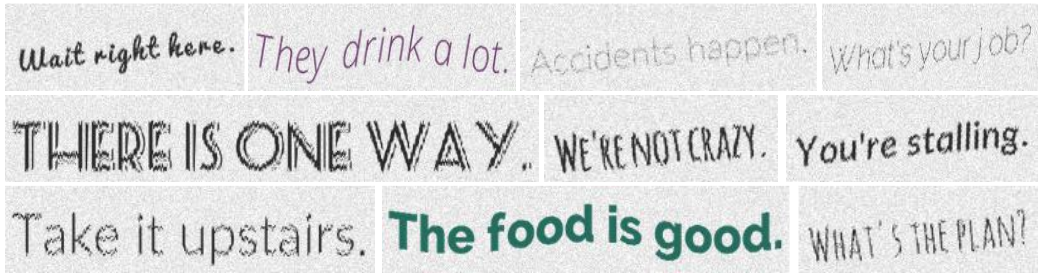}
    \caption{Data generated using TextRecognitionDataGenerator}
    \label{fig:textgenerator}
\end{figure}

Data augmentation was done to make the model better on misfit words like 'c0de' for the original word 'code'. To generate such augmented data, I used TextRecognitionDataGenerator \cite{textdatagen} to generate images for the English text in the translation pair. I used different configurations for all pairs of the words. I involved distortion, text color, fonts, blur, skew angle. I randomized the parameters for the generator and generated 125444 images for all the translation pairs available in the ANKI dataset for Spanish-English pairs. Some examples of the data images generated are shown in the figure \ref{fig:textgenerator}. The generated images were passed through the EasyOCR and the OCR predictions were stored as a dictionary with the keys set as the prediction and value as the original English text. Using the dictionary and the original translation pair, I mapped the OCR output to the corresponding Spanish translation. The pairs were appended into the \textbf{spa.txt} file (original spa.txt - translation pairs). 

\subsection{Models}
\begin{figure}[!h]
    \centering
    \includegraphics[width=\textwidth]{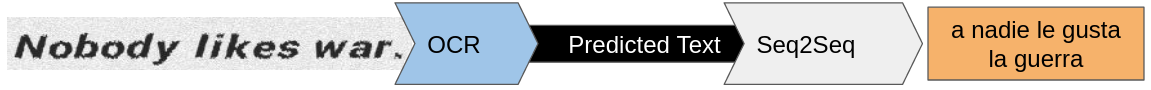}
    \caption{Flow of the complete TransDocs Script}
    \label{fig:flow}
\end{figure}

The pipeline of the project is split into two modules: OCR and Text language translation. The complete pipeline/flow can be seen in the figure \ref{fig:flow}. For the OCR module of the pipeline, I have explored pre-trained machine learning model namely, Tesseract \cite{tesseract} and EasyOCR \cite{easyocr}. I have tested out some examples on both the models. Tesseract and EasyOCR both give output as bounding box coordinates, predicted text and confidence level. Hence, testing out and comparing the two different models was fair as comparison was made using the same metric. I used the confidence level and the predicted text as the deciding metrices. Also, ease of integration, ease of installation on the system and prediction time was also taken into consideration. After performing some experimentation, I finalized EasyOCR as it gave higher accuracy in reading text in high distortion, noise, skew angle of the text and other kind of manipulations on the data. Also, it gives prediction at a very good rate once the model is loaded.

 The main focus of my project is to train a machine translation deep learning model\cite{uppala2023dynamic, kundu2018maharshi} based seq2seq architecture using LSTM. To understand better about the topic, I referred to some of the well known network like Google sequence to sequence \cite{seq2seqgoogle} which basically employs encoder to make a vector to represent the input and decoder, which basically converts back the embedding to a meaning sentence. The model involves using LSTM cells for the learning. These methods are proven to perform well and published by various research groups over the years. \cite{seq2seqgoogle} \cite{facebookseq}.
 
\begin{figure}[!h]
  \centering
  \includegraphics[width=\textwidth]{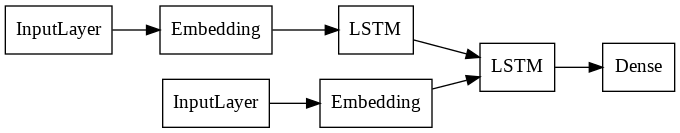}
  \caption{Model plot for the normal seq2seq model (without attention)}
  \label{fig:normalmodel}
\end{figure}

\begin{figure}[!h]
  \centering
  \includegraphics[width=\textwidth]{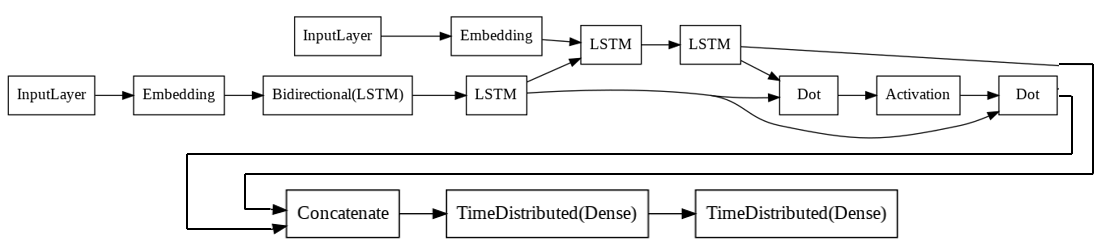}
  \caption{Model plot for the Attention seq2seq model (Luong et al. Attention)}
  \label{fig:attentionmodel}
\end{figure}

 In this project, I have built two seq2seq models as shown in the figure \ref{fig:normalmodel} and \ref{fig:attentionmodel}. The main difference with the two models is Luong et al. attention is incorporated to built upon the normal seq2seq model (figure \ref{fig:normalmodel}). A seq2seq model reads the source sentence using an encoder. An Encoder is used to to build a vector, a sequence of numbers that represents the sentence meaning. This can also be imagined as an encrypted format for the input sentence while maintaining the meaning of the sentence. Then a decoder is used to process the vector or encryption to give an output which is the predicted translation of the input sentence. This is often referred to as the encoder-decoder architecture. Such architecture can solve the machine translation very effectively as the LSTM cells/ memory cells used in these architecture are responsible for capturing the previously seen input and the current and try to make sense of the input. Making these architectures the best to perform machine translation. 

Taking advantage of the categorical nature of words, the model must be provided with the source and target embeddings to retrieve the corresponding word representations.To do that embeddings are required for both the source language (English) and target language (Spanish). I decided to learn the embeddings from scratch and used all the data pairs as I had a good amount of data. To create the embeddings, the first step is to select the vocabulary for each language.

In the model, the English sentences are sent into the input layer and sent into embedding layer to create english word embeddings. Then the vector generated using the embedding layer is sent into the Encoder in this case the LSTM layer. The LSTM layer generates a latent representation of the input sentence and generate an encrypted representation of the input sentence. Once the latent representation is ready, in-parallel other language (Spanish) is fed into the other input layer and continuted into the embeddings for the Spanish language. Given the embeddings for the Spanish database and the vector representation from the Encoder are fed into the Decoder. The decoded representation of the input is fed into a dense layer and the predictions are made. In this case, as the words are dealt as categorical, I have used Sparse Categorial Cross Entropy as loss function for the training purpose.

For the attention model, I planned on testing out the architecture with attention layer as I saw some publication on machine translation getting better results. To proceed with the architecture, I had to explore more on the attention blocks that can be used and understand those in-depth before building a model out of it. I explored Bahdanau et al. model \cite{bahdanau2014neural} and Luong et al. model \cite{Luong} attention blocks. There are pros and cons of each of the attention block. In Luong attention, it uses top hidden layer states in both of encoder and decoder. But Bahdanau attention simply takes concatenation of forward and backward source hidden state. Also, in Luong attention, the decoder hidden state at time, t is considered and attention score is calculated. From that we get the context vector which is concatenated with hidden state of the decoder and hence prediction is made. Whereas, in Bahdanu decoder hidden state at time, t-1 is considered, then the alignment is calculated and hence context vector is obtained. so, now the context vector is concatenated with the hidden state of the decoder at t-1. Hence, before it is sent for prediction, the concatenated vector goes into the LSTM/GRU cell. Also, how the score is calculated makes the two attention block quite different. In this project, I have opted to use Luong et al. with my attention model and the model plot can be seen in the figure \ref{fig:attentionmodel}. 

\section{Results}
\begin{figure}[!h]
    \centering
    \subfloat{{\includegraphics[width=0.4 \textwidth]{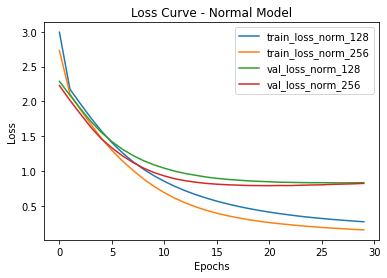}}}%
    \qquad
    \subfloat{{\includegraphics[width=0.4 \textwidth]{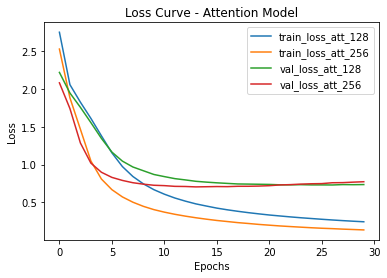}}}%
    \qquad
    \subfloat{{\includegraphics[width=0.4 \textwidth]{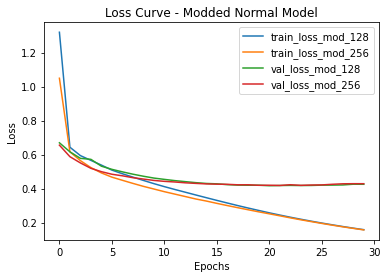}}}%
    \qquad
    \subfloat{{\includegraphics[width=0.4 \textwidth]{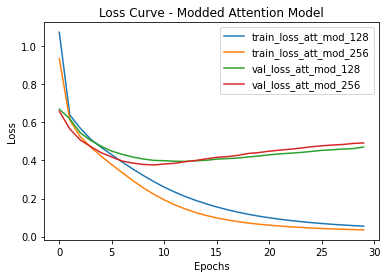}}}%
    \caption{Comparing loss curves for four configurations of trained models}%
    \label{fig:curves}%
\end{figure}
The idea I wanted to test in this project was successful. I wanted to build a model which could use even the misfit predictions from the OCR and still give me good language translations for the same. For example, there can be cases where the text has some skew angle due to the scanned document being tilted or not well-aligned while scanning. Also, there can be cases where the camera/scanner used to scan the document is not of high quality and/or there is some distortion in the image captured due to shakiness of the scanner, It can be hard for OCR to give a good prediction. Therefore, there can be outputs like, '+oday' for the word 'today', 'c0de' for the word 'code' and so on. For the same, I tested out the models described in the Models section above. I tested the normal model (figure \ref{fig:normalmodel}) and attention model (figure \ref{fig:attentionmodel}) with different configurations for the n\_units parameters for our hidden state. I have tested out different values for learning rate, dropout rate, English embedding dimension, Spanish embedding dimension and n\_units. I have finally chosen to stick with 0.001, 0.2, 256, 256 and [128, 256] for each parameter, respectively and opted adam as the optimizer.

I have trained both Normal model and Attention model with just the original eng-spa translation pairs with multiple configuration for n\_units but chose to show only 2 best performing parameter values [128, 256].  You can see the training and validation loss curves for normal model and attention curve for n\_units as 128 and 256 in the figure \ref{fig:curves} first row. We can see that for normal model, n\_units set as 256 gave lower validation loss as seen by the red curve in the first plot of the figure \ref{fig:curves}. In the attention model, we can see that n\_units set as 256 gives quite lower validation loss as seen in red curve in the second plot of the figure \ref{fig:curves}. We can see that the lowest validation loss is achieved way early as compared to n\_units as 128.

To improve the results, I did the data augmentation as described before. Doing that, I was able to generate misfit English - correct Spanish translation pairs to augment the dataset. Using the appended data, I trained the same models with multiple n\_units and the curves can be seen in the second row of the figure \ref{fig:curves}. We can see that 128 and 256 configurations gives very similar results for the normal model and also converge in very close number of epochs. Still the lowest among those configurations was 128 n\_units and can be seen in the red curve. For the attention model, we can see that when n\_units set as 256 out perform the 128 setup and also takes 3 less epochs to converge. 
\begin{figure}[!h]
    \centering
    \includegraphics[width = 0.5\textwidth]{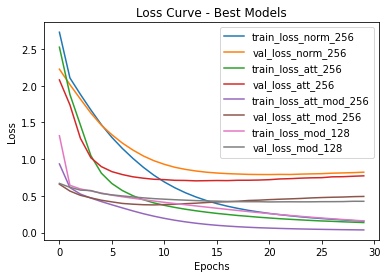}
    \caption{Loss Curve for best model from each type of models}
    \label{fig:compare_best}
\end{figure}

The best configurations from each of the model is plotted together to compare the performance of the augmentation and shown in the figure \ref{fig:compare_best}. We can see that the modded models i.e the models with augmentations outperformed the counter by a significant number and have resulted in a very good validation loss. You can find the validation loss for each of the model for both configurations and the epochs to achieve the best validation error are shown in the table \ref{tab:val_error}.

\begin{table}[!h]
\centering
\caption{Different models with lowest validation loss (SparseCategoricalCrossentropy) and epochs to achieve}
\label{tab:val_error}
\begin{tabular}{|l|l|l|}
\hline
\textbf{Model Varients} & \textbf{128 Units (Epochs)} & \textbf{256 Units (Epochs)} \\ \hline
Normal                  & 0.83004 (28)                & \textbf{0.79011 (21) }      \\ \hline
Attention               & 0.72957 (27)                & \textbf{0.70429 (14)}       \\ \hline
Modded Normal           & \textbf{0.41691 (21)}       & 0.41847 (22)                \\ \hline
Modded Attention        & 0.39486 (13)                & \textbf{0.37627 (10)}       \\ \hline
\end{tabular}
\end{table}

\begin{figure}[!h]
    \centering
    \includegraphics[width=0.8\textwidth]{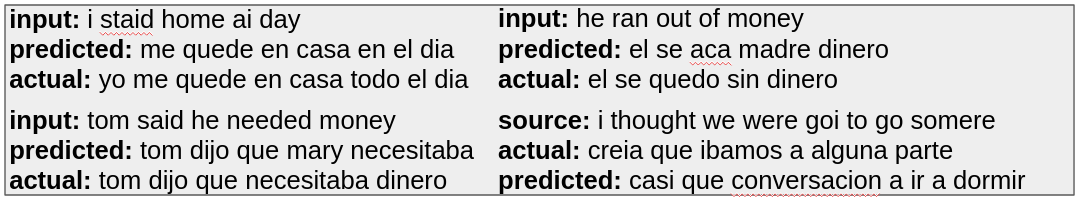}
    \caption{Predictions from the Attention Modded model}
    \label{fig:predictions}
\end{figure}

BLEU-4 is Bilingual Evaluation Understudy Score with uniform weights. BLEU-4 score is also evaluated on all the 4 model variants. Comparing my work with any of the benchmarking model like fairseq or Google seq2seq will be unfair as those models are trained on the billions of datapoints and with very high resources like fairseq is trained for 7 days on 8 parallel GPU. So, comparing my baseline with the modded is what I target for this project. In the table \ref{tab:bleu}, we can see that the Attention model works the best and gives the highest BLEU-4 score. This is because attention model is trained on more number of datapoints from the accuracte input sentences pool. This makes the model stronger in terms of BLEU score. Some predictions from the attention modded model are shown in the figure \ref{fig:predictions}

\begin{table}[!h]
\centering
\caption{BLEU-4 score for all model variants}
\label{tab:bleu}
\begin{tabular}{|l|l|}
\hline
\textbf{Model Variants} & \textbf{BLEU-4 Score} \\ \hline
Normal                  & 0.69578               \\ \hline
Attention               & \textbf{0.72013}      \\ \hline
Normal Modded           & 0.57910               \\ \hline
Attention Modded        & 0.65116               \\ \hline
\end{tabular}
\end{table}

\section{Discussion and Analysis}

Scanned documents are way more common and very few resources are available to actually deal with such documents effectively. There are technologies like Google Lens available which can perform text in wild translation and give some good results. Such a model is trained using billions and billions of data and takes over a week to train with multiple GPU setup. I believe there has to be some more research that has to go into the this topic. This topic holds a lot of potential as more and more smart devices are introduced everyday. I believe I have tried to introduce a novel pipeline to help customers. The work I have done is just a prototype and requires a lot of improvements, resources and ideation.

From the results, I have achieved during this project, I believe I have made a good pipeline at-least for one language pair. The results show that the model is able to perform well even on the test set which is basically taken from the erroneous/misfit predictions from the OCR. Also, the outputs shown in the figure \ref{fig:predictions} are very promising and can be improved further by augmenting the data even more.

In this project, I have taken some steps which can be build upon to get a better performing model. Because of limited resources, I have only augmented the data with one type of modification and can be repeated and hence billions of data points can be generated. In my case, I have produced 125444 augmented images and hence doubled the input dataset.

After all the in-depth study about the Seq2Seq models, I learnt that GRUs can be a better replacement for LSTM cells. This is because GRU can learn better with short sentences/words and hence better prediction from the model. This is helpful because I'm targeting word by word prediction.

\bibliography{ref.bib}{}

\begin{thebibliography}{10}

\bibitem{ANKI}
\url{http://www.manythings.org/anki/}.

\bibitem{textdatagen}
\url{https://github.com/Belval/TextRecognitionDataGenerator}.

\bibitem{easyocr}
\url{https://github.com/JaidedAI/EasyOCR}.

\bibitem{keras_official}
\url{https://blog.keras.io/a-ten-minute-introduction-to-sequence-to-sequence-learning-in-keras.html}.

\bibitem{DBLP:journals/corr/abs-1904-01906}
Jeonghun Baek, Geewook Kim, Junyeop Lee, Sungrae Park, Dongyoon Han, Sangdoo
  Yun, Seong~Joon Oh, and Hwalsuk Lee.
\newblock What is wrong with scene text recognition model comparisons? dataset
  and model analysis.
\newblock {\em CoRR}, abs/1904.01906, 2019.

\bibitem{bahdanau2014neural}
Dzmitry Bahdanau, Kyunghyun Cho, and Yoshua Bengio.
\newblock Neural machine translation by jointly learning to align and
  translate.
\newblock {\em arXiv preprint arXiv:1409.0473}, 2014.

\bibitem{facebookseq}
Jonas Gehring, Michael Auli, David Grangier, Denis Yarats, and Yann~N. Dauphin.
\newblock Convolutional sequence to sequence learning.
\newblock {\em CoRR}, abs/1705.03122, 2017.

\bibitem{kar2023keydetect}
Soumyatattwa Kar, Abhishek Bamotra, Bhavya Duvvuri, and Radhika Mohanan.
\newblock Keydetect --detection of anomalies and user based on keystroke
  dynamics, 2023.

\bibitem{8543868}
Neeraj Kumar, Phanikrishna Uppala, Karthik Duddu, Hari Sreedhar, Vishal Varma,
  Grace Guzman, Michael Walsh, and Amit Sethi.
\newblock Hyperspectral tissue image segmentation using semi-supervised nmf and
  hierarchical clustering.
\newblock {\em IEEE Transactions on Medical Imaging}, 38(5):1304--1313, 2019.

\bibitem{kundu2018maharshi}
Jogendra~Nath Kundu.
\newblock Maharshi gor, phani krishna uppala, and r venkatesh babu.
  unsupervised feature learning of human actions as trajectories in pose
  embedding manifold.
\newblock In {\em IEEE Winter Conference on Applications of Computer Vision
  (WACV)}, 2018.

\bibitem{Luong}
Minh{-}Thang Luong, Hieu Pham, and Christopher~D. Manning.
\newblock Effective approaches to attention-based neural machine translation.
\newblock {\em CoRR}, abs/1508.04025, 2015.

\bibitem{tesseract}
Ray Smith.
\newblock An overview of the tesseract ocr engine.
\newblock In {\em Proc. Ninth Int. Conference on Document Analysis and
  Recognition (ICDAR)}, pages 629--633, 2007.

\bibitem{seq2seqgoogle}
Ilya Sutskever, Oriol Vinyals, and Quoc~V Le.
\newblock Sequence to sequence learning with neural networks.
\newblock In {\em Advances in neural information processing systems}, pages
  3104--3112, 2014.

\bibitem{Uppala-2021-129108}
Phani~Krishna Uppala.
\newblock Exemplar-free video retrieval.
\newblock Master's thesis, Carnegie Mellon University, Pittsburgh, PA, August
  2021.

\bibitem{uppala2023dynamic}
Phani~Krishna Uppala, Abhishek Bamotra, and Raj Kolamuri.
\newblock Dynamic object removal for effective slam, 2023.

\bibitem{uppala2023learning}
Phani~Krishna Uppala, Abhishek Bamotra, Shriti Priya, and Vaidehi Joshi.
\newblock Learning video embedding space with natural language supervision,
  2023.

\bibitem{vaswani2017attention}
Ashish Vaswani, Noam Shazeer, Niki Parmar, Jakob Uszkoreit, Llion Jones,
  Aidan~N Gomez, {\L}ukasz Kaiser, and Illia Polosukhin.
\newblock Attention is all you need.
\newblock In {\em Advances in neural information processing systems}, pages
  5998--6008, 2017.

\end{thebibliography}
\bibliographystyle{plain}
\vspace{5cm}
\fbox{Any code for the project are made available on \href{https://github.com/abhishekbamotra/TransDocs}{https://github.com/abhishekbamotra/TransDocs}}
\end{document}